\documentclass[10pt,twocolumn,letterpaper]{article}

\usepackage{wacv}              

\definecolor{wacvblue}{rgb}{0.21,0.49,0.74}
\usepackage[pagebackref,breaklinks,colorlinks,allcolors=wacvblue]{hyperref}
\usepackage[ruled,vlined]{algorithm2e} 
\usepackage[table]{xcolor} 
\SetKwInput{KwIn}{Input}
\SetKwInput{KwOut}{Output}
\usepackage[accsupp]{axessibility}

\def\wacvPaperID{3028} 
\def\confName{WACV}
\def\confYear{2026}

\title{
MorphXAI: An Explainable Framework for Morphological Analysis of Parasites in Blood Smear Images}

\author{
Aqsa Yousaf$^{1}$ \qquad
Sint Sint Win$^{2}$ \qquad
Megan Coffee$^{2}$ \qquad
Habeeb Olufowobi$^{1}$ \\[0.2cm]
$^{1}$Department of Computer Science and Engineering, University of Texas at Arlington, USA \\
$^{2}$Department of Medicine, Division of Infectious Diseases, NYU Grossman School of Medicine, USA \\[0.15cm]
{\tt\small aqsa.yousaf@uta.edu, habeeb.olufowobi@uta.edu} \\
{\tt\small sintsint.win@nyulangone.org, megan.coffee@nyulangone.org}
}

\begin{document}
\maketitle

\begingroup
\renewcommand\thefootnote{}
\footnotetext{The research reported in this paper was supported by AIM-AHEAD Coordinating Center, award number OTA-21-017, and was, in part, funded by the National Institutes of Health Agreement No. 1OT2OD032581.}
\addtocounter{footnote}{0}
\endgroup

\begin{abstract}
Parasitic infections remain a pressing global health challenge, particularly in low-resource settings where diagnosis still depends on labor-intensive manual inspection of blood smears and the availability of expert domain knowledge.~While deep learning models have shown strong performance in automating parasite detection, their clinical usefulness is constrained by limited interpretability.~Existing explainability methods are largely restricted to visual heatmaps or attention maps, which highlight regions of interest but fail to capture the morphological traits that clinicians rely on for diagnosis.~In this work, we present MorphXAI, an explainable framework that unifies parasite detection with fine-grained morphological analysis.~MorphXAI integrates morphological supervision directly into the prediction pipeline, enabling the model to localize parasites while simultaneously characterizing clinically relevant attributes such as shape, curvature, visible dot count, flagellum presence, and developmental stage.~To support this task, we curate a clinician-annotated dataset of three parasite species (Leishmania, Trypanosoma brucei, and Trypanosoma cruzi) with detailed morphological labels, establishing a new benchmark for interpretable parasite analysis.~Experimental results show that MorphXAI not only improves detection performance over the baseline but also provides structured, biologically meaningful explanations. 

\end{abstract}

\begin{figure}[t]
    \centering
    \includegraphics[width=0.95\linewidth]{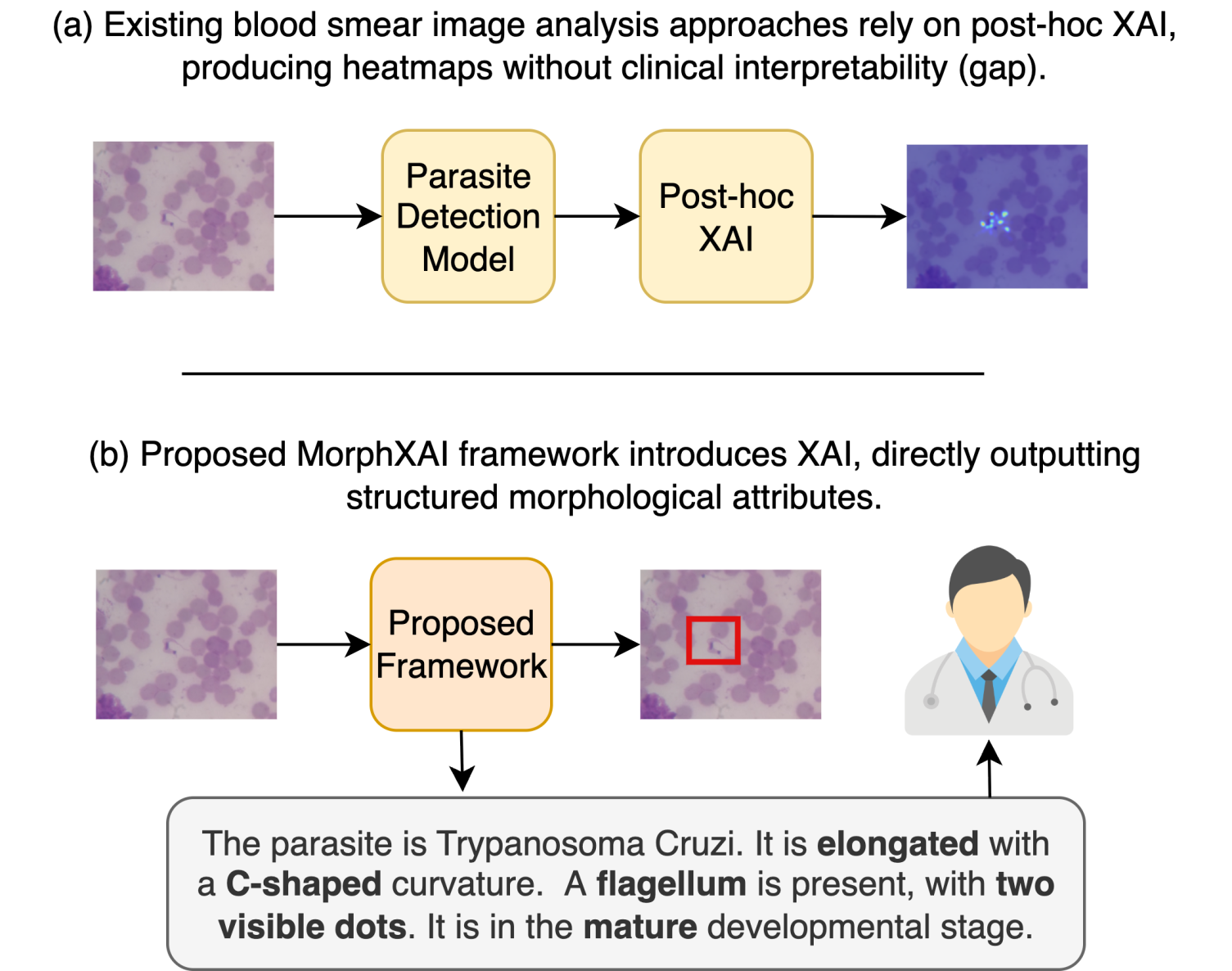}
    \caption{(a) Existing approaches rely on post-hoc XAI, producing heatmaps without clinical interpretability (gap). 
    (b) Our proposed framework introduces an XAI method that directly outputs structured morphological attributes, interpretable by clinicians.}
    \label{fig:intro_obxai}
    \vspace{-0.3cm}
\end{figure}
  
\section{Introduction}
\label{sec:intro}

Parasitic infections remain a major global health threat, causing substantial morbidity and mortality each year~\cite{scanes2018parasites}. Diagnosis typically relies on microscopic examination of stained blood smears, which is accurate but labor-intensive.~The challenge is especially severe in low-resource regions, where timely diagnosis is hindered by a shortage of trained microscopists~\cite{bradbury2022have}. 


Automated analysis of blood smears using deep learning offers a promising alternative~\cite{tekle2024deepleish}.~Recent automated parasite localization studies increasingly used CNN-based detectors such as Faster R-CNN, SSD, and YOLO, which have demonstrated strong performance on microscopy images~\cite{liu2023aidman}.~Transformer-based detectors further improve robustness for small objects, complex morphologies, and noisy backgrounds~\cite{ruiz2025dt4peis, shehzadi2025object}. These advances show that accurate detection is feasible.~Nevertheless, detection alone does not meet clinical requirements.~Clinicians base their decisions not only on parasite presence but also on morphological traits which are essential for species differentiation and treatment planning~\cite{wong2014molecular}.~Existing detection models do not capture this reasoning. 


To address the interpretability gap, researchers have applied the available explainability methods to parasite detection.~Techniques such as Grad-CAM, Grad-CAM++, and SHAP provide visual explanation by highlighting image regions influencing a prediction~\cite{jillahi2024semantic}, while transformer-based approaches often rely on attention maps~\cite{islam2022explainable,yousaf2025beyond}.~These methods offer useful visualizations, but they have three major limitations.~First, they are post-hoc and may not fully reflect the features the model relied on during prediction.~Second, they largely duplicate localization signals already conveyed by bounding boxes.~Third, they fail to capture the morphological cues that guide expert decisions.~As a result, current explainability techniques fall short of the interpretability required for medical deployment as explained in Figure~\ref{fig:intro_obxai}.

In this work, we introduce MorphXAI, an explainable framework for parasite analysis that integrates morphological reasoning directly into the detection pipeline.~Instead of relying on post-hoc visualization, MorphXAI predicts clinically relevant attributes 
alongside parasite localization. This design ensures that each detection is accompanied by structured, biologically meaningful explanations aligned with clinical practice.~To support training and evaluation, we curate a clinician-annotated dataset of \textit{Leishmania}, \textit{Trypanosoma brucei}, and \textit{Trypanosoma cruzi} with detailed morphological labels, establishing a new benchmark for interpretable parasite analysis.~Experiments demonstrate that MorphXAI preserves strong detection performance while transforming black-box predictions into interpretable diagnostic reports, bridging the gap between automated detection and clinician-style reasoning.

\noindent Our main contributions are summarized as follows:
\begin{itemize}
\item We curate a clinician-annotated dataset of three parasite species with detailed morphological labels, establishing a new benchmark for interpretable parasite analysis in blood smears.
\item We propose MorphXAI, the first explainable framework for blood parasite analysis that integrates parasite detection with fine-grained morphological characterization, producing outputs aligned with clinical reasoning.
\item We introduce a morphology-guided supervision scheme that jointly optimizes detection and morphological prediction, ensuring that the learned explanations remain structured and biologically meaningful.
\item We demonstrate that integrating domain-specific morphological supervision into detection can serve as a generalizable strategy for bridging the gap between visual explanations and expert reasoning in biomedical vision tasks.
\item We conduct extensive experiments and ablation studies to validate the framework, demonstrating that MorphXAI preserves strong detection accuracy while providing rich, interpretable explanations.
\end{itemize}

\section{Related Work}
\label{sec:related_work}

\begin{figure*}[h!]
    \centering
    \includegraphics[width=0.95\textwidth]{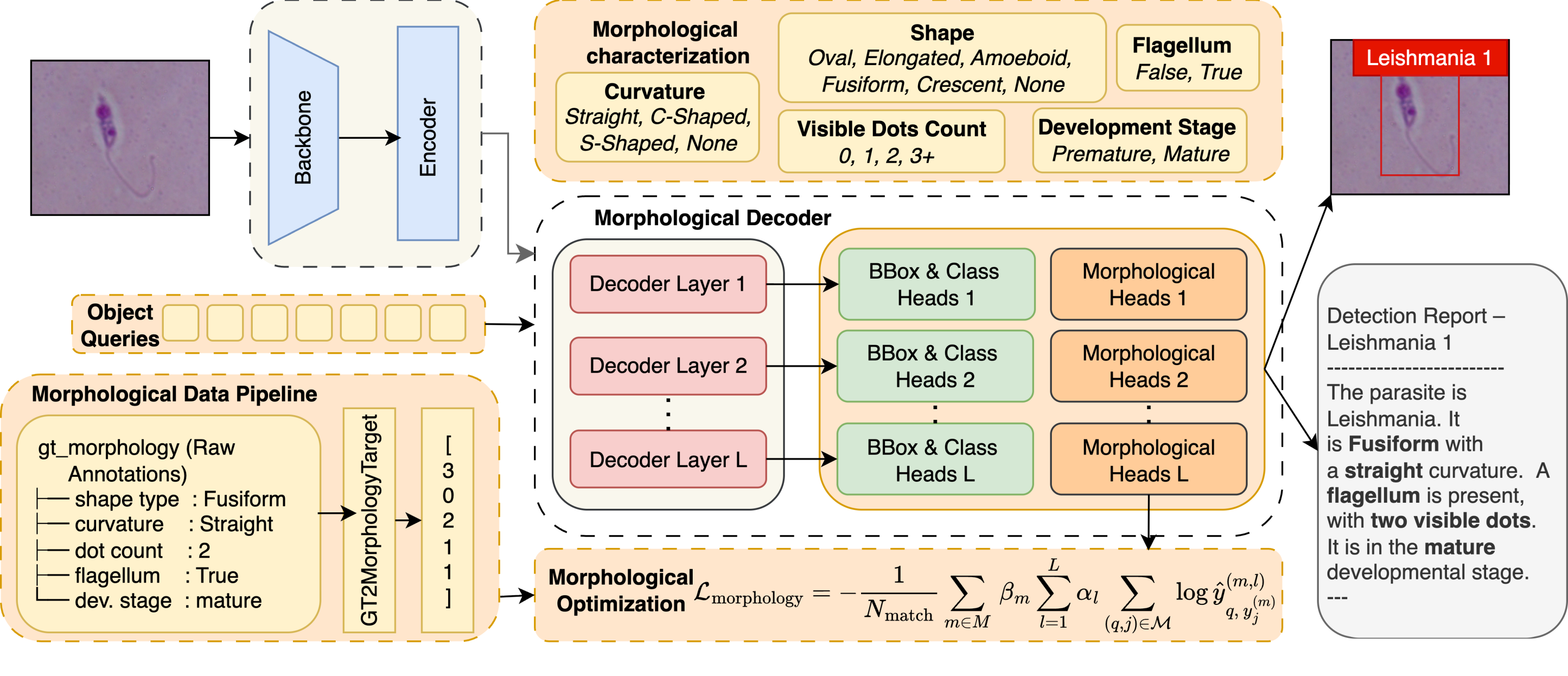}
    \caption{Overview of MorphXAI. The framework introduces a morphological decoder that jointly predicts bounding boxes, parasite class, and clinically relevant attributes. A dedicated data pipeline converts clinician annotations into supervision targets, and a morphology-aware optimization guides training. The model produces both detections and structured morphological reports, enabling interpretable parasite analysis.}
    \label{fig:morphology_framework}
\end{figure*}

Automated analysis of blood smear images has become an active research area, driven by the need for scalable and reliable tools to support clinical diagnosis of parasitic infections.~Deep learning models, particularly CNNs and more recently transformers, have demonstrated strong performance in parasite detection and classification tasks. While these approaches demonstrate impressive detection accuracy, they primarily treat parasite recognition as an object detection problem, overlooking the morphological traits that underpin clinical diagnosis.

\subsection{CNN-Based Parasite Detection}
CNN-based architectures have been widely explored for parasite detection in blood smears.~Early work focused on adapting general-purpose detectors such as Faster R-CNN, SSD, and YOLO to parasitology tasks.~For example, Tekle et al.~\cite{tekle2024deepleish} evaluated multiple CNN detectors for Leishmania detection, reporting the best performance with YOLOv5.~Several studies have since introduced tailored improvements to YOLO variants.~Sukumarran et al.~\cite{sukumarran2024optimised} optimized YOLOv4 through layer pruning and backbone replacement for malaria parasite detection, while Zedda et al.~\cite{zedda2023yolo} proposed YOLO-PAM, a parasite-attention model built on YOLOv8.~Similarly, Mura et al.~\cite{mura2025yolo} developed YOLO-Tryppa, incorporating a dedicated small-object head and ghost convolutions to detect parasites.~Liu et al.~\cite{liu2023aidman} combined YOLOv5 for cell detection with a transformer-based “attentional aligner” for parasite classification.~Other efforts emphasize deployment in constrained environments.~For instance, Lin et al.~\cite{lin2025artificial} presented a smartphone-based diagnostic system running lightweight CNN detectors such as SSD-MobileNetV2 and YOLOv8 for real-time inference, and Gonçalves et al.~\cite{gonccalves2023detection} addressed visceral leishmaniasis through segmentation of bone marrow smears using CNN-based models. 

Collectively, these CNN-based approaches demonstrate strong detection accuracy and versatility, but their predictions are restricted to bounding boxes or class probabilities, offering little biological context for clinical interpretation.

\subsection{Transformer-Based Parasite Detection}
More recently, transformer architectures have been explored to address CNN limitations~\cite{shehzadiefficient,shehzadi2025object}, particularly in handling small objects, morphological diversity, and noisy imaging conditions.~Ruiz-Santaquiteria et al.~\cite{ruiz2025dt4peis} introduced DT4PEIS, combining a DETR-based detector with the Segment Anything Model for parasitic egg segmentation.~Nakarmi et al.~\cite{nakarmi2023deep} benchmarked Deformable DETR against Faster R-CNN and YOLOv8 for detecting Giardia and Cryptosporidium, showing that transformers remain competitive even on low-quality smartphone images.~In blood smears, Guemas et al.~\cite{guemas2023automatic} applied RT-DETR to classify multiple Plasmodium species. 

Despite their advantages, transformer-based approaches likewise limit their outputs to bounding boxes or categorical labels, without reasoning about the morphological traits that clinicians rely on for species differentiation.


\subsection{Explainability in Parasite Detection Models}
Beyond detection accuracy, another line of work has sought to improve model transparency and trust through explainability techniques.~Gradient-based methods such as Grad-CAM~\cite{selvaraju2017grad}, Grad-CAM++~\cite{chattopadhay2018grad}, and SHAP~\cite{lundberg2017unified} generate heatmaps that highlight salient regions, and have been applied in several CNN frameworks.~For example, SPCNN~\cite{ahamed2025improving}, LeishFuNet~\cite{sadeghi2024deep}, and capsule network variants~\cite{alawfi2025hybrid} used these techniques to confirm that predictions focused on parasitized cells.~Other studies~\cite{Patel_Vahora_2025, jillahi2024semantic} combined CNN outputs with model-agnostic tools like LIME~\cite{ribeiro2016should} or incorporated ontology-based reasoning for more structured explanations. Transformer-based approaches follow similar trends, relying on Grad-CAM or attention maps to visualize focus regions~\cite{yousaf2025beyond}.

While these efforts improve transparency, they remain constrained in three ways.~First, they are post-hoc methods applied after training, raising concerns about faithfulness to the model’s reasoning.~Second, they are primarily visual, producing heatmaps or attention maps that overlap with bounding boxes rather than describing meaningful traits.~Third, they are not aligned with clinical practice, since they do not describe morphological descriptors that guide expert diagnosis. 

Thus, existing explainability methods increase interpretability but they fail to capture clinically salient morphological descriptors such as nucleus-to-cytoplasm ratio, flagellum presence, or cell shape irregularities.~Bridging this gap requires models that do not merely highlight pixels, but articulate morphological traits aligned with the diagnostic reasoning.

\section{Methodology}

\begin{table*}[t]
\centering
\resizebox{\textwidth}{!}{%
\begin{tabular}{lcl}
\toprule
\rowcolor{orange!20}
\textbf{Morphological Attribute} & \textbf{Mapping $f_a$} & \textbf{Values $\mapsto$ Indices} \\
\midrule
Shape  & $M_{\text{shape}} \to \{0\!-\!5\}$ & oval $\mapsto 0$, elongated $\mapsto 1$, amoeboid $\mapsto 2$, fusiform $\mapsto 3$, crescent $\mapsto 4$, other $\mapsto 5$ \\
Curvature  & $M_{\text{curv}} \to \{0\!-\!3\}$ & straight $\mapsto 0$, C-shaped $\mapsto 1$, S-shaped $\mapsto 2$, round $\mapsto 3$ \\
Dot count  & $M_{\text{dot}} \to \{0\!-\!3\}$ & 0 $\mapsto 0$, 1 $\mapsto 1$, 2 $\mapsto 2$, 3+ $\mapsto 3$ \\
Flagellum  & $M_{\text{flag}} \to \{0,1\}$& False $\mapsto 0$, True $\mapsto 1$ \\
Development stage & $M_{\text{stage}} \to \{0,1\}$ & immature $\mapsto 0$, mature $\mapsto 1$ \\
\bottomrule
\end{tabular}}
\caption{Mapping of morphological attributes to class indices.}
\label{tab:morphology_mapping}
\end{table*}

We propose \textbf{MorphXAI}, an explainable detection framework that unifies parasite localization with fine-grained morphological characterization. Unlike conventional detectors that output only bounding boxes and class labels, MorphXAI produces a structured ``explanation vector” for each detection, consisting of clinically meaningful traits.~These attributes are directly predicted during detection rather than inferred post-hoc, ensuring that explanations are faithful to the features actually used for localization. Figure~\ref{fig:morphology_framework} provides an overview.

MorphXAI builds on the real-time transformer detector RT-DETRv3~\cite{wang2409rt}, chosen for its efficiency and accuracy. Our key contribution lies in integrating \emph{morphological reasoning} directly into the detection process.~The framework consists of three components:  
(i) a \textbf{morphological decoder} that predicts multiple morphological attributes alongside bounding boxes and parasite species,  
(ii) a \textbf{structured data pipeline} that converts expert annotations into multi-attribute supervision targets, and  
(iii) a \textbf{joint optimization strategy} that couples detection with attribute prediction to enforce biologically meaningful explanations.


\subsection{Morphological Decoder}
The morphological decoder extends the RT-DETR architecture into a multi-task setting.~Standard decoders predict bounding boxes and class logits; our decoder introduces parallel morphology heads that predict clinically relevant traits. This design ensures that detection and explanation are tightly coupled, since both rely on the same query features.

The decoder extends the transformer architecture with structured prediction modules,~where each module addresses specific morphological attribute. These modules operate alongside standard box and class heads and produce attribute-specific outputs for every query. Importantly, the same query features are shared between detection and morphological prediction, which binds the two tasks together and ensures that attribute predictions are directly grounded in the evidence used for localization.

Formally, let $H \in \mathbb{R}^{N \times Q \times d}$ denote the sequence of hidden states from the transformer decoder, with $N$ layers, $Q$ object queries, and hidden dimension $d=256$. For each decoder layer $i \in \{1,\dots,N\}$, the hidden state $H^{(i)} \in \mathbb{R}^{Q \times d}$ feeds into three sets of heads:  

\[
\begin{aligned}
\hat{y}_{\text{box}}^{(i)} &= \mathrm{MLP}_{\text{box}}^{(i)}(H^{(i)}) \in \mathbb{R}^{Q \times 4}, \\
\hat{y}_{\text{cls}}^{(i)} &= Linear_{\text{cls}}^{(i)} H^{(i)} \in \mathbb{R}^{Q \times C_{\text{det}}}, \\
\hat{y}_{m}^{(i)} &= Linear_{m}^{(i)} H^{(i)}  \in \mathbb{R}^{Q \times C_m}, \quad m \in \mathcal{M}.
\end{aligned}
\]




Here, $\mathcal{M}$ is the set of morphological attributes (\textit{shape type}, \textit{curvature}, \textit{dot count}, \textit{flagellum presence}, and \textit{developmental stage}), and $C_m$ is the class space for attribute $m$. 

Each decoder layer contains one bounding box head, one classification head, and $|\mathcal{M}|$ morphology heads.~Each decoder layer contributes predictions for all tasks, leading to $N \times |\mathcal{M}|$ morphology predictions per sample, all supervised during training.~This layer-wise deep supervision ensures that attribute learning is reinforced at every stage of decoding rather than confined to the final layer. During inference, only the outputs from the last decoder layer are retained, forming the structured explanation vector associated with each parasite detection. 

\vspace{2mm}
\noindent
\textbf{Advantages of the Design.} 
This framework reformulates parasite detection as a multi-task setting where each bounding box is accompanied not only by a species label but also by a structured set of morphological attributes.~By reframing parasite detection as structured multi-task prediction, the morphological decoder provides three advantages:  
\begin{itemize}
    \item  it explicitly disentangles features for clinically significant attributes,  
    \item  it enables independent optimization of attributes with different convergence behaviors, and  
    \item it supports modular extensibility, allowing new traits to be added without redesigning the detection backbone.
\end{itemize}

\begin{figure}[t]
    \centering
    \includegraphics[width=\linewidth]{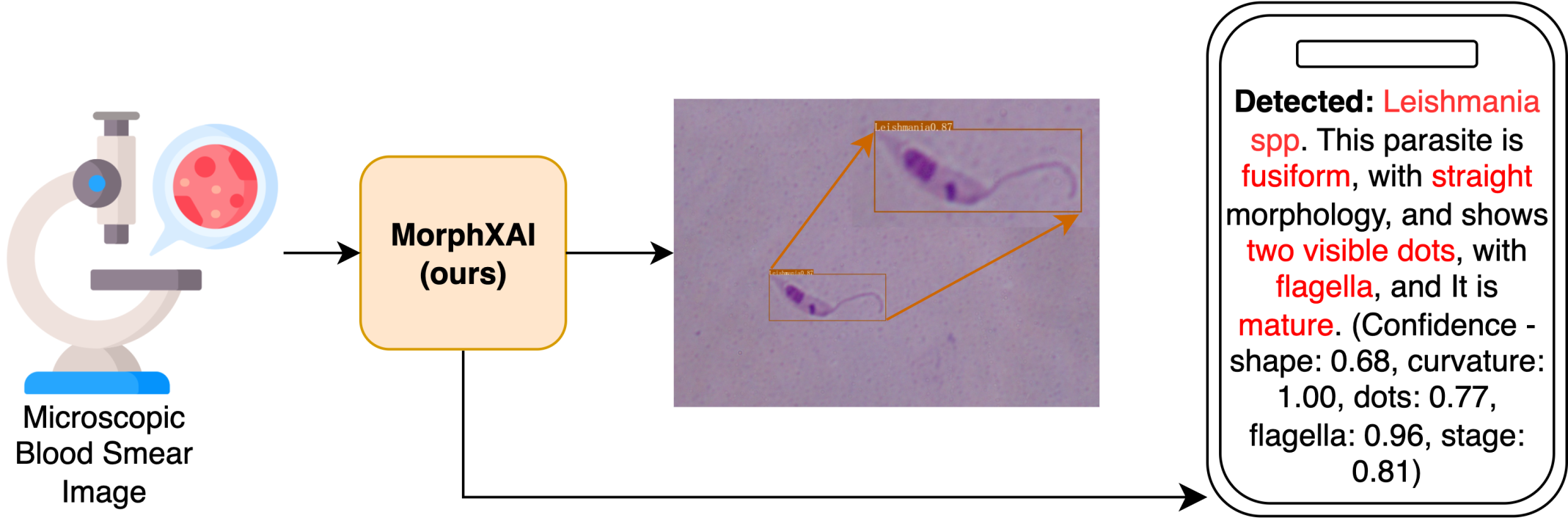}
    \caption{Inference pipeline of MorphXAI. Given a microscopic blood smear image, our framework detects parasites and augments each prediction with structured, clinically meaningful explanations.~The output includes parasite type, morphological traits, and confidence scores, providing interpretable results aligned with clinical reasoning.}
    \label{fig:inference_pipeline}
\end{figure}

Beyond numerical predictions,~MorphXAI assembles outputs into concise textual reports summarizing each parasite in terms of its morphology (Figure~\ref{fig:inference_pipeline}). This bridges the gap between black-box predictions and the clinician-style reasoning required in practice.



\subsection{Morphological Data Pipeline}
To couple parasite detection with morphological reasoning, we design a data pipeline that converts expert annotations into structured supervision targets for multi-task training, ensuring consistency and clinical validity.~Each parasite instance in the dataset is annotated by clinicians with traits such as shape, curvature, dot count, flagellum presence, and developmental stage.~These annotations must be converted into machine-readable class indices while preserving semantic consistency.~We implement this via the GT2MorphologyTarget module, which maps each attribute value to an integer index through a predefined function $f_m$ while preserving semantic consistency.~The mapped attributes ensure consistent biological interpretation across the dataset.~The morphological ground truth (GT) data is organized as a list of dictionaries, where each entry corresponds to a parasite instance annotated with attributes. For example, an annotated instance may be represented as:  

\begin{verbatim}
gt_morphology = [
  { 'shape_type': 'elongated',
    'curvature': 'C-shaped',
    'dot_count': 2,
    'flagellum_present': True,
    'development_stage': 'mature' }
]
\end{verbatim}

Formally, Let $\mathcal{M}$ denote the set of attributes and $M_m$ the possible values for attribute $m$. For each $m \in \mathcal{M}$, we define a mapping function
\[
f_m : M_m \rightarrow \{0, 1, \dots, |M_m| - 1\}, \quad 
F = \{ f_m \mid m \in \mathcal{M} \}.
\]

The complete mapping is shown in Table~\ref{tab:morphology_mapping}. For image $i$ with $N_i$ parasites, the annotation for instance $j$ is written as
\[
gt_{(i,j)} = \{ gt^m_{(i,j)} \mid m \in \mathcal{M} \},
\]
and converted into class indices as
\[
\hat{y}_{(i,j)} = \{ f_m(gt^m_{(i,j)}) \mid m \in \mathcal{M} \}.
\]

Across a batch of $B$ images, this produces attribute-specific supervision tensors
\[
T_m \in \mathbb{Z}^{N_{\text{total}} \times 1}, \quad 
N_{\text{total}} = \sum_{i=1}^B N_i,
\]
which are aligned with decoder queries via Hungarian matching, providing a multi-task supervision signal that couples parasite localization with clinically meaningful morphological traits, ensuring that explanations remain aligned with expert reasoning.

\subsection{Morphological Optimization}
We design a joint optimization strategy that couples parasite detection with morphology prediction, ensuring that explanatory traits are learned alongside localization.~We supervise morphology at three levels: per-attribute, per-layer, and across attributes. These are then combined with the detection loss into the total objective.

\vspace{2mm}
\noindent
\textbf{Per-attribute:} Each attribute $m \in \mathcal{M}$, is trained with cross-entropy loss over matched query–parasite pairs:

\[
\mathcal{L}_{\text{morph},m}
= - \frac{1}{N_{\text{match}}} 
\sum_{(q,j) \in \mathcal{H}} 
\log \hat{y}^{(m)}_{q,\, y^{(m)}_j},
\]
where $\mathcal{H}$ denotes the set of query--ground truth matches from Hungarian matching, $q$ indexes a query, $j$ the corresponding ground-truth parasite, $y^{(m)}_j$ the class index for attribute $m$, and $N_{\text{match}}$ the number of matched pairs. 

\vspace{2mm}
\noindent
\textbf{Per-layer:} Since morphology heads are attached to every decoder layer, each attribute contributes a loss at each layer $l \in \{1,\dots,L\}$. The per-attribute loss is aggregated across layers:
\[
\mathcal{L}_{\text{morph},m}^{\text{total}} = \sum_{l=1}^{L} \alpha_l \, \mathcal{L}_{\text{morph},m}^{(l)},
\]
with uniform layer weights $\alpha_l$ in our setup. 

\vspace{2mm}
\noindent
\textbf{Across attributes:} The overall morphology supervision is then obtained by summing across all attributes
\[
\mathcal{L}_{\text{morphology}} = \sum_{m \in \mathcal{M}} \mathcal{L}_{\text{morph},m}^{\text{total}}.
\]

The complete training loss integrates detection supervision with morphological supervision:
\[
\mathcal{L}_{\text{total}} = \mathcal{L}_{\text{det}} + \lambda \, \mathcal{L}_{\text{morphology}},
\]
where $\lambda$ balances the contribution of morphology supervision.~During training, the outputs from all decoder layers are supervised to guide learning.~During inference, only the predictions from the final decoder layer are used for both detection and morphological attributes.

Detection supervision combines classification, bounding box regression, and denoising terms:
\[
\mathcal{L}_{\text{det}} 
= \mathcal{L}_{\text{cls}} 
+ \mathcal{L}_{\text{bbox}} 
+ \mathcal{L}_{\text{dn}},
\]
where $\mathcal{L}_{\text{cls}}$ is a cross-entropy classification loss, $\mathcal{L}_{\text{bbox}}$ combines $\ell_1$ and GIoU penalties for localization, and $\mathcal{L}_{\text{dn}}$ regularizes denoising queries that stabilize Hungarian matching.

This joint optimization ensures that each query learns not only to localize parasites but also to explain them through structured, clinically interpretable traits, providing faithful transparency while preserving strong detection accuracy.





\begin{figure*}[t]
    \centering
    \begin{subfigure}[t]{0.45\textwidth}
        \centering
        \includegraphics[width=\textwidth]{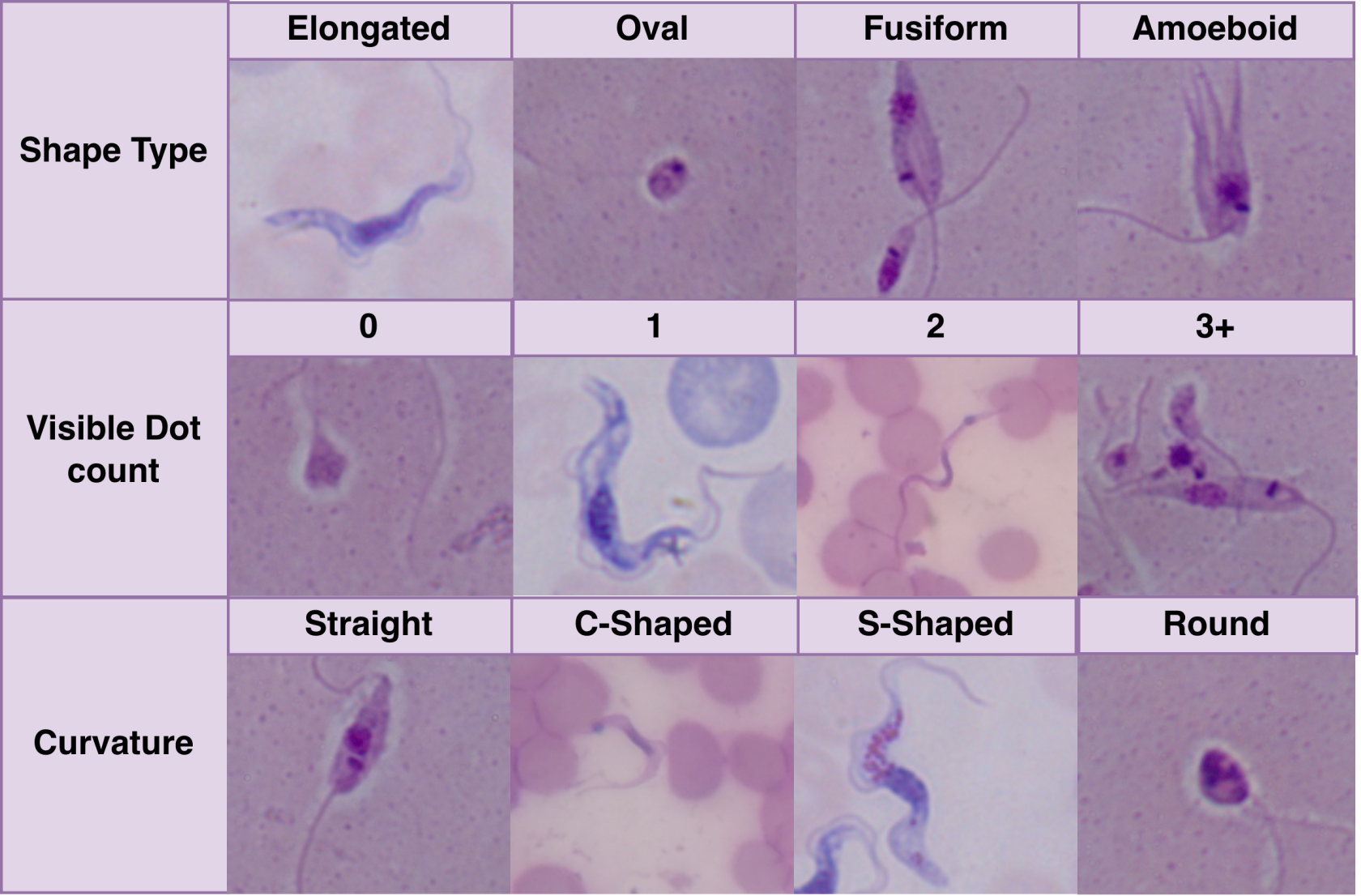}
        \caption{Examples of (i) \textbf{Shape type} (elongated, oval, fusiform, amoeboid), 
        (ii) \textbf{Visible dot count} (0, 1, 2, or 3+ dots), 
        and (iii) \textbf{Curvature} (straight, C-shaped, S-shaped, or round).}
    \end{subfigure}\qquad
    \begin{subfigure}[t]{0.39\textwidth}
        \centering
        \includegraphics[width=\textwidth]{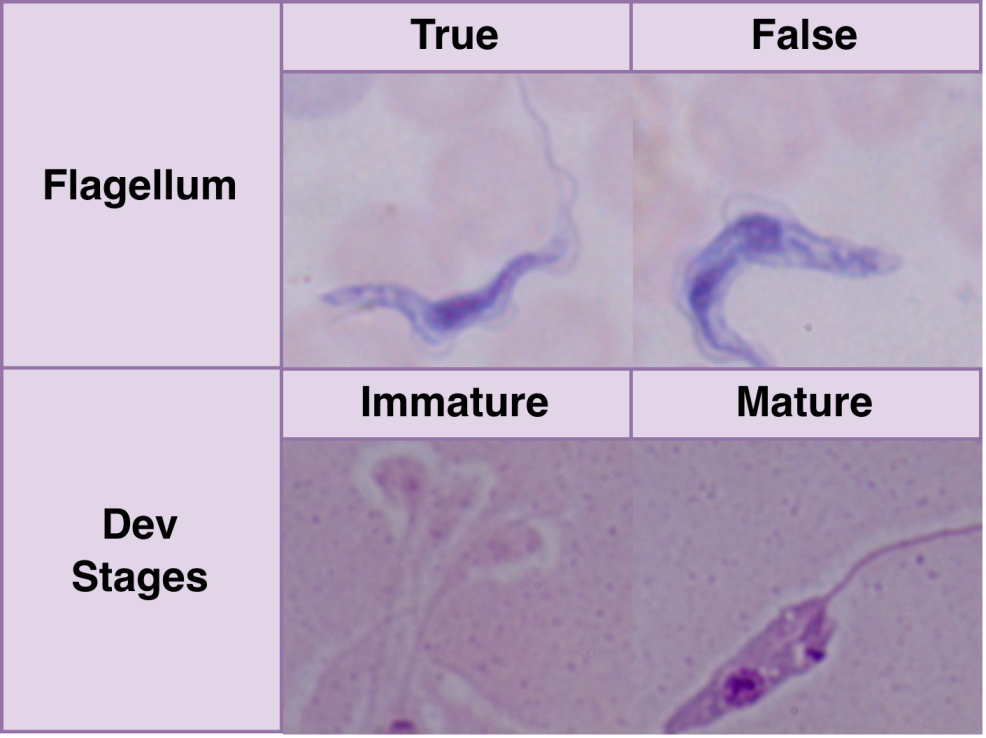}
        \caption{Examples of (iv) \textbf{Flagellum presence} (present/absent) and 
        (v) \textbf{Developmental stage} (immature/mature).}
    \end{subfigure}
    \caption{Clinician-verified morphological attributes used in our dataset. 
    These structured attributes capture the key cues that form the basis for our explainability framework.}
    \label{fig:morph_attributes}
    \vspace{-0.3cm}
\end{figure*}

\section{Experiments}
\label{sec:Exp}

\subsection{Dataset}
We curated a clinician-annotated dataset of 1,818 high-resolution blood smear images containing 5,903 parasite instances of Leishmania, Trypanosoma cruzi, and Trypanosoma brucei, which are causative agents of WHO-priority neglected tropical diseases~\cite{WHO_NTDs_2025}. Raw images were sourced from two public datasets~\cite{morais2022automatic, jiang2021parasites}, while all annotations, bounding boxes and morphological attributes, were provided by an expert clinician. Figure~\ref{fig:morph_attributes} illustrates examples of the morphological attributes annotated by the clinician. The dataset captures real-world variability with multi-scale objects, overlapping parasites, and diverse staining conditions.~Images are high resolution (mean: 1768×1449 pixels), preserving fine morphological cues essential for species differentiation. Detailed dataset statistics, including class distribution and split sizes, are provided in Table~\ref{tab:dataset}.

\begin{table}[t]
\centering
\caption{Statistics of the proposed parasite dataset, including image counts, per-class annotations, and split distribution.}
\label{tab:dataset}
\begin{tabular}{lcc}
\toprule
\rowcolor{orange!20}
\textbf{Instances} & \textbf{Train} & \textbf{Validation} \\
\midrule
Images & 1,523 & 295 \\
\textit{Leishmania} & 1,558 & 393 \\
\textit{T. cruzi} & 2,028 & 385 \\
\textit{T. brucei} & 1,376 & 163 \\
\midrule
\textbf{Total} & 4,962 & 941 \\
\bottomrule
\end{tabular}
\vspace{-0.3cm}
\end{table}

\subsection{Implementations}
All experiments were conducted on an NVIDIA RTX A6000 GPU with 48 GB VRAM. Our framework, MorphXAI, introduces a new morphological explainable decoder with integrated optimization that jointly predicts bounding boxes, parasite classes, and clinically relevant attributes. This decoder replaces the standard decoder in RT-DETRv3.~Unless otherwise specified, we adopted the default hyperparameters from RT-DETRv3 to ensure a fair comparison.~The batch size was fixed at 12 across all experiments due to memory constraints.~Training was performed for 50 epochs.~We evaluated our model using a ResNet-18 backbone to emphasize computational efficiency and clinical deployability in resource-constrained settings, while additional experiments with ResNet-34 are presented in the ablation study.~The morphology loss weight was set to $\lambda = 0.5$, with uniform layer weights $\alpha = 1.0$ across decoder layers.
\section{Results}
\label{sec:results}

\begin{table}[t]
\centering
\caption{Comparison of detection performance between RT-DETRv3 and MorphXAI on the validation set.~MorphXAI achieves slightly higher AP while additionally producing structured morphological explanations.}
\label{tab:detection}
\resizebox{\linewidth}{!}{%
\begin{tabular}{lcccc}
\toprule
\rowcolor{orange!20}
\textbf{Model} & \textbf{AP$^{.50:.95}$} & \textbf{AP$^{.50}$} & \textbf{AP$^{.75}$} & \textbf{AR$^{.50:.95}$} \\
\midrule
RT-DETRv3~\cite{wang2409rt} & 47.3 & 86.1 & 44.7 & \textbf{65.7} \\
MorphXAI (ours) & \textbf{48.2} & \textbf{87.5} & \textbf{45.1} & 64.2 \\
\bottomrule
\end{tabular}}
\end{table}

Table~\ref{tab:detection} compares the baseline detector with MorphXAI. While both models achieve similar accuracy, MorphXAI slightly improves overall detection performance (0.482 AP$^{.50:.95}$ vs. 0.473).~More importantly, it augments each detection with fine-grained morphological attributes, transforming the output from bounding boxes alone into structured, clinically interpretable results as shown in Figure~\ref{fig:inference_pipeline}. This reframing moves automated analysis beyond recognition accuracy toward explanations that reflect expert diagnostic reasoning.

Table~\ref{tab:morphology} shows that MorphXAI reliably predicts morphological traits that clinicians use in diagnosis.~Morphology classification is evaluated in a detection-conditioned manner.~For each predicted bounding box with IoU $\geq 0.5$ against ground truth, we assess whether the associated morphological attributes are correctly predicted.~This ensures that interpretability is measured only on correctly localized parasites, mirroring realistic diagnostic practice where explanations are meaningful only after successful detection.

\begin{figure}[t]
    \centering
    \includegraphics[width=0.99\linewidth]{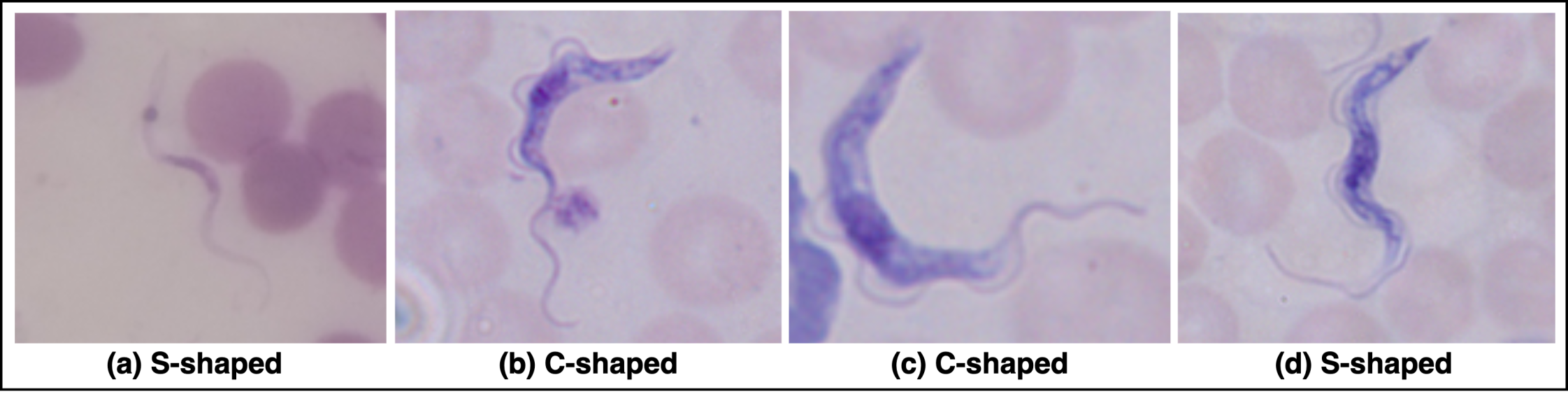}
    \caption{ 
    Morphological attribute, Curvature, was defined strictly on the parasite body.     Clinician annotators labeled the curvature attribute based only on the parasite body, without considering the orientation of the free-floating flagellum.
    In images (b) and (c), the body is clearly C-shaped, but in (b) the flagellum extends outward in an S-like curve. 
    This makes it visually similar to (a), where the parasite body itself is S-shaped. 
    Such cases illustrate the biological complexity of curvature assessment. These visual overlaps explain why curvature prediction accuracy is relatively low. With larger datasets and refined attribute definitions, this challenge can be further overcome.
    }
    \label{fig:curvature_ambiguity}
\end{figure}

\begin{figure*}[t]
    \centering
    \includegraphics[width=\linewidth]{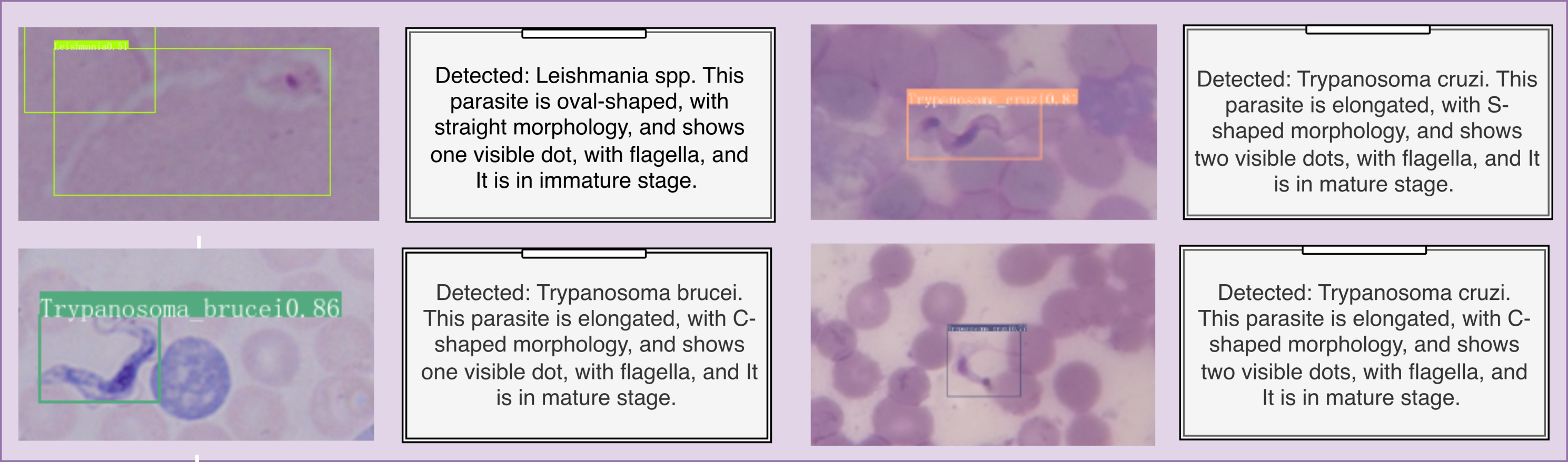}
    \caption{Qualitative results from MorphXAI. For each detected parasite, the model outputs both the bounding box and a structured explanation describing its morphological traits. These explanations provide interpretable context beyond bounding-box predictions, aligning automated outputs with the reasoning clinicians use in practice.}
    \label{fig:qualitative_results}
\end{figure*}

Attributes with clear visual boundaries, such as flagellum presence (96.6\%) and developmental stage (93.9\%), are recognized with high reliability. Shape type (88.6\%) and visible chromatin dots (85.7\%) are also predicted consistently. Curvature is comparatively more difficult, with an accuracy of 61.7\%. This is partly due to annotation ambiguity. As explained in Figure~\ref{fig:curvature_ambiguity}, in many cases, the parasite body follows a C-shaped curve, but a free-floating flagellum extends outward and produces an S-like outline. Such variations complicate labeling and make curvature prediction inherently less stable.~Despite this challenge, MorphXAI captures meaningful signals across all traits, and the structured outputs provide built-in explanations that reflect features familiar to clinical experts, as shown in Figure~\ref{fig:qualitative_results}.

\begin{table}[t]
\centering
\caption{Morphology classification accuracy of MorphXAI (ours) on the validation set. 
Accuracy is computed in a detection-conditioned manner (IoU $\geq 0.5$), ensuring attribute predictions are evaluated only on correctly localized parasites.}

\label{tab:morphology}
\begin{tabular}{lc}
\toprule
\rowcolor{orange!20}
\textbf{Morphological Attribute} & \textbf{Accuracy (\%)} \\
\midrule
Shape           & 88.6 \\
Curvature           & 61.7 \\
Dot count           & 85.7 \\
Flagellum presence  & 96.6 \\
Development Stage            & 93.9 \\
\bottomrule
\end{tabular}
\end{table}

\begin{table}[t]
\centering
\caption{Inference efficiency of MorphXAI (ours) compared to RT-DETRv3 on the validation set. 
MorphXAI adds attribute prediction heads to the decoder but maintains real-time performance.}
\label{tab:efficiency}
\begin{tabular}{lcc}
\toprule
\rowcolor{orange!20}
\textbf{Model} & \textbf{FPS} & \textbf{Time / Image (ms)} \\
\midrule
RT-DETRv3~\cite{wang2409rt}    & 58.62 & 17.1 \\
MorphXAI & 42.73 & 23.4 \\
\bottomrule
\end{tabular}
\end{table}

Table~\ref{tab:efficiency} presents inference efficiency.~MorphXAI processes images at 42.7 FPS (23.4 ms per image), compared to 58.6 FPS for the baseline.~The slight increase in runtime is expected, since the decoder now produces both detection and attribute predictions. However, the system still operates comfortably in real-time, and the additional interpretability justifies the small computational overhead. This demonstrates that clinically meaningful explainability can be embedded into the detection pipeline without sacrificing deployability.

\section{Ablation Studies}
We first examine the effect of the morphology loss weight $\lambda$ (Table~\ref{tab:ablation_lambda} and Table~\ref{tab:ablation_lambda_morph}). A moderate setting of $\lambda=0.5$ achieves the best overall balance, yielding the highest detection accuracy (48.2 AP$^{.50:.95}$ and 45.1 AP$^{.75}$). Smaller values, such as $\lambda=0.2$, reduce the contribution of morphology supervision and slightly weaken both detection and attribute prediction. Increasing $\lambda$ beyond 0.5 does not improve overall AP, but provides marginal gains for certain morphological attributes.~However, these improvements come at the cost of reduced detection precision, making $\lambda=0.5$ the most effective trade-off.

Across all values of $\lambda$, attributes with clear visual boundaries, such as flagellum presence ($\geq 96\%$) and developmental stage ($\geq 92\%$), remain highly reliable.~In contrast, subtle cues like curvature show greater sensitivity to the loss of weight, improving from 61.7\% at $\lambda=0.2$ to 67.5\% at $\lambda=2.0$. 

\begin{table}[t]
\centering
\caption{Effect of morphology loss weight ($\lambda$) on detection performance. 
Results are reported on the validation set.}
\label{tab:ablation_lambda}
\begin{tabular}{lcccc}
\toprule
\rowcolor{orange!20}
\textbf{\textbf{$\lambda$}} & \textbf{AP$^{.50:.95}$} & \textbf{AP$^{.50}$} & \textbf{AP$^{.75}$} & \textbf{AR$^{.50:.95}$} \\
\midrule
0.2 & 46.2 & 87.4 & 41.6 & 62.5 \\
0.5 & \textbf{48.2} & 87.5 & \textbf{45.1} & 64.2 \\
1.0 & 46.9 & 87.1 & 43.6 & \textbf{65.6} \\
2.0 & 46.9 & \textbf{87.7} & 43.6 & 63.7 \\
\bottomrule
\end{tabular}
\end{table}

\begin{table}[t]
\centering
\caption{Effect of morphology loss weight ($\lambda$) on morphology classification accuracy (\%). 
Accuracy is computed only on correctly localized parasites (IoU $\geq 0.5$).}
\label{tab:ablation_lambda_morph}
\resizebox{\linewidth}{!}{%
\begin{tabular}{lccccc}
\toprule
\rowcolor{orange!20}
\textbf{$\lambda$} & \textbf{Shape} & \textbf{Curvature} & \textbf{Dot count} & \textbf{Flagellum} & \textbf{Dev. Stage}  \\
\midrule
0.2 & 88.2 & 61.7 & 87.0 & \textbf{97.2} & 93.9  \\
0.5 & \textbf{88.6} & 61.7 & 85.7 & 96.6 & 93.9  \\
1.0 & 88.1 & 62.9 & 86.5 & 97.1 & 92.1  \\
2.0 & 88.1 & \textbf{67.5} & \textbf{88.1} & 96.2 & \textbf{94.1}  \\
\bottomrule
\end{tabular}}
\end{table}

\begin{table}[t]
\centering
\caption{Effect of backbone choice on detection performance with $\lambda=0.5$. 
Results are reported for ResNet-18 and ResNet-34 on the validation set.}
\label{tab:backbone}
\begin{tabular}{lcccc}
\toprule
\rowcolor{orange!20}
\textbf{Backbone} & \textbf{AP$^{.50:.95}$} & \textbf{AP$^{.50}$} & \textbf{AP$^{.75}$} & \textbf{AR$^{.50:.95}$} \\
\midrule
ResNet-18 & \textbf{48.2} & \textbf{87.5} & \textbf{45.1} & \textbf{64.2} \\
ResNet-34 & 45.0 & 84.3 & 41.6 & 63.6 \\
\bottomrule
\end{tabular}
\end{table}

We evaluate the impact of backbone choice by comparing ResNet-18 and ResNet-34 with $\lambda=0.5$ (Table~\ref{tab:backbone}). ResNet-18 outperforms ResNet-34 in detection performance. This result reflects the relatively modest size of our clinician-annotated dataset: deeper backbones such as ResNet-34 are more prone to overfitting, leading to weaker generalization on the validation set. By contrast, the lighter ResNet-18 strikes a better balance between capacity and regularization, resulting in higher AP.

Morphological attribute classification (Table~\ref{tab:backbone-class}), however, is slightly stronger with ResNet-34.~Since attribute accuracy is measured only on correctly localized parasites ($\text{IoU} \geq 0.5$), this suggests that although ResNet-34 detects fewer parasites overall, the ones it does localize, support more reliable morphology predictions.~ResNet-18 maximizes detection coverage, while ResNet-34 yields marginally stronger morphology characterization on a smaller subset of confident detections. This trade-off highlights the suitability of ResNet-18 for deployment scenarios where recall is critical, while also showing that deeper backbones can enhance the robustness of morphological explanations.



\section{Limitations and Future Work}
MorphXAI demonstrates the feasibility of embedding morphological explainability into real-time parasite detection, but certain limitations remain. The dataset is clinician-annotated but limited in both size and diversity, covering only three parasite species. 
In addition, the framework currently models five morphological attributes, while clinical diagnosis often relies on a broader range of morphological cues. Future work will focus on curating larger datasets that span additional parasite species and morphological variations, enabling the use of richer attribute sets. As the number of attributes increases, maintaining separate prediction heads for each attribute may become impractical for real-time deployment. A promising direction is the design of a unified prediction head, implemented as a deeper or multi-layer module, that can jointly model all attributes while preserving efficiency.~This would extend the scalability of MorphXAI to capture a broader range of morphological characteristics without compromising real-time performance.

\begin{table}[t]
\centering
\caption{Effect of backbone choice on morphology classification accuracy (\%). 
Results are reported on the validation set and computed only on correctly localized parasites (IoU $\geq 0.5$).}
\label{tab:backbone-class}
\begin{tabular}{lcc}
\toprule
\rowcolor{orange!20}
\textbf{Morphological Attribute} & \textbf{Resnet18} & \textbf{Resnet34} \\
\midrule
Shape type         & \textbf{88.6} & 87.9 \\
Curvature          & 61.7 & \textbf{63.5} \\
Dot count          & 85.7 & \textbf{87.0} \\
Flagellum presence & 96.6 & \textbf{96.9} \\
Development Stage  & 93.9 & \textbf{94.1}   \\  
\bottomrule
\end{tabular}
\vspace{-0.6cm}
\end{table}
\section{Conclusion}
We introduced MorphXAI, a framework that unifies parasite detection with structured morphological characterization, advancing explainability beyond post-hoc visualization. The framework jointly detects parasites and predicts clinically relevant attributes such as shape, curvature, chromatin dots, flagellum presence, and developmental stage, producing outputs that align with the diagnostic reasoning used by clinicians. To enable this task, we curated a clinician-annotated dataset of three parasite species with detailed morphological labels, establishing a new benchmark for interpretable parasite analysis.~Experiments demonstrate that MorphXAI preserves strong detection accuracy while providing reliable, real-time explanations. 





{
    \small
    \bibliographystyle{ieeenat_fullname}
    \bibliography{main}
}

\end{document}